\documentclass[sigconf]{acmart}

\usepackage{makecell}
\usepackage{tabularx}
\usepackage{adjustbox}
\usepackage{graphicx}
\usepackage{subcaption}
\usepackage{amsmath}
\AtBeginDocument{%
  }

\copyrightyear{2025} 
\acmYear{2025} 
\setcopyright{acmlicensed}
\acmConference[ICMR '25]{Proceedings of the 2025
International Conference on Multimedia Retrieval}{June 30-July 3,
2025}{Chicago, IL, USA}
\acmBooktitle{Proceedings of the 2025 International Conference on Multimedia
Retrieval (ICMR '25), June 30-July 3, 2025, Chicago, IL, USA}
\acmDOI{10.1145/3731715.3733283}
\acmISBN{979-8-4007-1877-9/2025/06}

\newcommand{\systemname}{{BRepFormer}}
\begin{document}

%%
%% The "title" command has an optional parameter,
%% allowing the author to define a "short title" to be used in page headers.
\title{\systemname: Transformer-Based B-rep Geometric Feature Recognition}

%%
%% The "author" command and its associated commands are used to define
%% the authors and their affiliations.
%% Of note is the shared affiliation of the first two authors, and the
%% "authornote" and "authornotemark" commands
%% used to denote shared contribution to the research.

\author{Yongkang Dai}
\orcid{0009-0000-5848-9433}
\affiliation{%
  \institution{School of Software, \\Northwestern Polytechnical University}
  \city{Xi'an}
  % \state{Shaanxi}
  \country{China}
}
\email{daiyongkang@mail.nwpu.edu.cn}

\author{Xiaoshui Huang}
\orcid{0000-0002-3579-538X}
\affiliation{%
  \institution{School of Public Health, \\Shanghai Jiaotong \\ University}
  \city{Shang hai}
  \country{China}
}
\email{huangxiaoshui@163.com}

\author{Yunpeng Bai}
\orcid{0009-0008-7578-0079}
\affiliation{%
  \institution{School of Computing,\\ National University  of \\ Singapore}
  \country{Singapore}
}
\email{bai_yunpeng99@u.nus.edu}

\author{Hao Guo}
\orcid{0009-0008-5091-0738}
\affiliation{%
  \institution{School of Software, \\Northwestern Polytechnical University}
  \city{Xi'an}
  % \state{Shaanxi}
  \country{China}
}
\email{guoh0215@mail.nwpu.edu.cn}

\author{Hongping Gan}
\orcid{0000-0002-4853-5077}
\affiliation{%
  \institution{School of Software, \\Northwestern Polytechnical University}
  \city{Xi'an}
  % \state{Shaanxi}
  \country{China}
}
\email{ganhongping@nwpu.edu.cn}

\author{Ling Yang}
\orcid{0009-0002-1263-3879}
\affiliation{%
  \institution{Hangzhou Global Science and Technology Innovation Center, Zhejiang University }
  \city{Hang zhou}
  % \state{Zhejiang}
  \country{China}
}
\email{3110101176@zju.edu.cn}

\author{Yilei Shi}
\authornote{The corresponding author.}
\orcid{0000-0001-7386-0026}
\affiliation{%
  \institution{School of Software, \\Northwestern Polytechnical University}
  \city{Xi'an}
  % \state{Shaanxi}
  \country{China}
}
\email{yilei_shi@nwpu.edu.cn}

%%
%% By default, the full list of authors will be used in the page
%% headers. Often, this list is too long, and will overlap
%% other information printed in the page headers. This command allows
%% the author to define a more concise list
%% of authors' names for this purpose.
\renewcommand{\shortauthors}{Yongkang Dai et al.}

%%
%% The abstract is a short summary of the work to be presented in the
%% article.
\begin{abstract}
% In the field of modern intelligent manufacturing, Machining Feature Recognition (MFR) serves as a crucial link between Computer-Aided Design (CAD) and Computer-Aided Manufacturing (CAM). However, traditional MFR methods often fall short in effectively capturing the intricate topological and geometric characteristics of complex CAD models. To address this challenge, we propose a novel deep learning-based MFR network, \systemname. Our approach, while processing the features of CAD models, meticulously extracts and encodes the geometric and topological attributes of the models, thereby attaining a profound comprehension of the model characteristics. Additionally, we introduce a novel Transformer structure for information propagation of the encoded features. During each iteration of the network, we incorporate a bias that combines edge features and face features to reinforce geometric constraints on each face, thereby enhancing the model's performance. We have also designed a complex CAD dataset, Complex B-rep Feature Dataset (CBF), which is more complex and closely aligned with industrial applications compared to traditional datasets. It comprises 1,500 CAD models in B-rep format. Experimental results demonstrate that \systemname achieves superior performance on the MFInstSeg, MFTRCAD, and CBF datasets. 

%The recognition of geometric features on B-rep models is an important research direction in the field of intelligent manufacturing.
Recognizing geometric features on B-rep models is a cornerstone technique for multimedia content-based retrieval and has been widely applied in intelligent manufacturing. 
However, previous research often merely focused on Machining Feature Recognition (MFR), falling short in effectively capturing the intricate topological and geometric characteristics of complex geometry features. In this paper, we propose \systemname, a novel transformer-based model to recognize both machining feature and complex CAD models' features. \systemname~ encodes and fuses the geometric and topological features of the models. Afterwards, \systemname~ utilizes a transformer architecture for feature propagation and a recognition head to identify geometry features. During each iteration of the transformer, we incorporate a bias that combines edge features and topology features to reinforce geometric constraints on each face. In addition, we also proposed a dataset named Complex B-rep Feature Dataset (CBF), comprising 20,000 B-rep models. By covering more complex B-rep models, it is better aligned with industrial applications. The experimental results demonstrate that \systemname~ achieves state-of-the-art accuracy on the MFInstSeg, MFTRCAD, and our CBF datasets. 
\end{abstract}

%%
%% The code below is generated by the tool at http://dl.acm.org/ccs.cfm.
%% Please copy and paste the code instead of the example below.
%%



%%
%% Keywords. The author(s) should pick words that accurately describe
%% the work being presented. Separate the keywords with commas.
%% A "teaser" image appears between the author and affiliation
%% information and the body of the document, and typically spans the
%% page.
% \begin{teaserfigure}
%   \includegraphics[width=\textwidth]{sampleteaser}
%   \caption{Seattle Mariners at Spring Training, 2010.}
%   \Description{Enjoying the baseball game from the third-base
%   seats. Ichiro Suzuki preparing to bat.}
%   \label{fig:teaser}
% \end{teaserfigure}

% \received{20 February 2007}
% \received[revised]{12 March 2009}
% \received[accepted]{5 June 2009}

%%
%% This command processes the author and affiliation and title
%% information and builds the first part of the formatted document.
\maketitle

%%%%%%%%% BODY TEXT
\section{Introduction}
\label{sec:intro}

Geometric feature recognition serves as a critical link between Computer-Aided Design (CAD) and Computer-Aided Manufacturing (CAM). It is a cornerstone technique for multimedia content-based retrieval and plays a key role in automating manufacturing processes, improving efficiency, and reducing human errors. 
While traditional rule-based geometric feature recognition methods are widely used in industry, they are labor-intensive and struggle to adapt to complex geometric variations and topological changes \cite{henderson1984computer, joshi1988graph, vandenbrande1993spatial}. 
To address these limitations, learning-based approaches have been introduced, often converting CAD models into intermediate representations such as point clouds \cite{lei2022mfpointnet}, voxels \cite{ning2023part}, or images \cite{2Dmanufacturing}. 
However, these transformations lead to key topological and geometric information loss, increased computational costs, and reduced recognition accuracy. Although recent deep learning models that directly process boundary representation (B-rep) data have shown promise in preserving geometric and topological details, as well as improving recognition accuracy~\cite{cao2020graph}, challenges remain in handling highly complex topologies and diverse manufacturing processes, limiting their practical effectiveness. 

% To address the limitations of existing methods, we propose a novel deep learning-based machining feature recognition network, \systemname, which leverages the Transformer architecture to enhance feature recognition capabilities. Our approach, while processing the boundary representation (B-rep) of CAD models, comprehensively extracts and encodes the geometric features and global topological features of the CAD models, thereby achieving a profound comprehension of the CAD models. Specifically, our method incorporates a novel Transformer structure to facilitate information propagation within the network. During each iteration, the network enforces constraints on edges and faces, thereby enhancing the performance of feature recognition. In this manner, BrepFormer not only resolves the issue of insufficient utilization of geometric and topological information in existing methods but also significantly improves the accuracy of machining feature recognition through its innovative network design. 

% <For all designs, like the graph-based transformer, you need to give reasons. For e.g., we model it as a graph structure because graph structure could reserve xx and enhance xx, but the reason is still unclear, why is that?>

% Second paragraph: following the mentioned gap, present our model, give some intuition why our proposed method (architecture) could fill the gap.

We propose \systemname, a transformer-based geometric feature recognition network that leverages a transformer architecture to effectively capture and process B-rep features. 
Unlike previous approaches that suffer from information loss and limited topological awareness \cite{ma2019automatic}, \systemname~directly operates on the B-rep, ensuring high-fidelity feature extraction. 
Specifically, our model extracts both the geometric and topological features of the CAD model from multiple perspectives. During the feature encoding stage, geometric edge features and topological features are processed by separate encoders and then fused to form an attention bias, which serves as a constraint input into the transformer module. 
Meanwhile, the extracted geometric face features are encoded into tokens together with a virtual face. These tokens serve as carriers of information that are fed into the transformer module, facilitating deep interaction and information fusion among the features. 
Finally, the recognition head accepts the output from the transformer module and fuses the global and local features within it, achieving high-precision recognition of geometric features. 
% Our model consists of four key components: (1) a Feature Extraction Module, which comprehensively extracts geometric and topological features; (2) a Feature Encoder Module, which transforms raw inputs into a structured representation optimized for learning; (3) a Transformer Block, which applies a Transformer structure with an attention bias constraint to enhance feature interactions; and (4) a Recognition Head, which fuses the extracted features and classifies B-rep faces.
By integrating these components, \systemname~effectively propagates information across the CAD model structure, improving geometric feature recognition accuracy while preserving critical geometric and topological relationships. 
% 看看要不要删 model overview，如果删就得多写transformer
% 对topo geom分别采样编码，融合 (以保留xx information details)，将token ... 形成bias --> serial tokens --> as constraints --> transformer.

% Results
% Extensive experimental validations were conducted on the MFInstSeg, MFTRCAD, and CBF datasets in this study, with results demonstrating that \systemname~ achieved state-of-the-art recognition accuracy across all tests. Particularly when dealing with complex and diverse geometric shapes as well as various types of machining features, \systemname~ exhibited remarkable capabilities. For instance, on the MFTRCAD dataset, the recognition accuracy of \systemname~ reached 93.16\%, representing an improvement of approximately 3.28 percentage points compared to the previous best method. Moreover, our approach provided a detailed analysis of the impact of different components on overall performance through ablation studies, further substantiating the effectiveness and superiority of the proposed method. Specifically, it was found that the incorporation of global constraints led to enhanced performance in handling complex feature interactions. 
We conducted experiments on the public MFInstSeg~\cite{wu2024aagnet} and MFTRCAD~\cite{MFTRCAD} datasets, as well as our CBF dataset. The results demonstrate that \systemname~achieves state-of-the-art recognition accuracy on these three datasets. Notably, \systemname~excels in recognizing complex geometric structures and diverse machining features. On the MFTRCAD \cite{MFTRCAD} dataset, it achieved a 93.16\% recognition accuracy, surpassing the previous best method by 3.28 percentage points. To further validate the effectiveness of our approach, we performed detailed ablation studies, analyzing the impact of each model component on overall performance. The results shows that every feature component helps boost the model's accuracy. 
%The ablation study results indicate that each feature component contributes to an improvement in the model's accuracy. 

% In summary, the \systemname~ proposed in this study not only addresses the issue of information loss present in existing methods but also significantly improves the performance of machining feature recognition through innovative network design, thereby laying a solid foundation for further development in the field of intelligent manufacturing. The main contributions of this paper can be summarized as follows: 
In summary, our proposed network, \systemname, effectively leverages the inherent information within the B-rep structure and achieves high-precision geometric feature recognition performance. The key contributions of this work are as follows: 

\begin{itemize}
    % \item Our method comprehensively extracts the geometric features and global topological features of CAD models, thereby achieving a profound understanding of the CAD models. 
    % \item We design a novel Transformer network architecture that incorporates edge and face constraints during the information propagation process, thereby enhancing the accuracy of feature recognition. 
    % \item We propose a complex CAD dataset tailored for industrial applications, filling the gap in the availability of complex CAD model datasets. 
    % \item Empirical validations on multiple benchmark datasets show that \systemname~ has reached the most advanced recognition accuracy and performs well when dealing with complex and variable geometric shapes. 
    \item We develop a method (Sec. \ref{featureExtractionModel} and \ref{featureEncoderModel}) that effectively extracts and encodes both topological and geometric features from CAD models, enabling a more informative and structured representation for machine learning models. 
    \item We introduce a novel transformer-based architecture (Sec. \ref{transformerModule}) that enforces edge and face constraints during information propagation, significantly improving feature recognition accuracy. 
    \item We introduce a CAD dataset (Sec. \ref{datasetCBF}) that is more aligned with industrial applications, offering a more complex collection of CAD models to contribute to the geometric feature recognition. 
    \item Our \systemname~ model achieves state-of-the-art accuracy on the experimental datasets, demonstrating the superiority of our approach. 
\end{itemize}

\begin{figure*}[t]
    \centering
    \includegraphics[width=\textwidth]{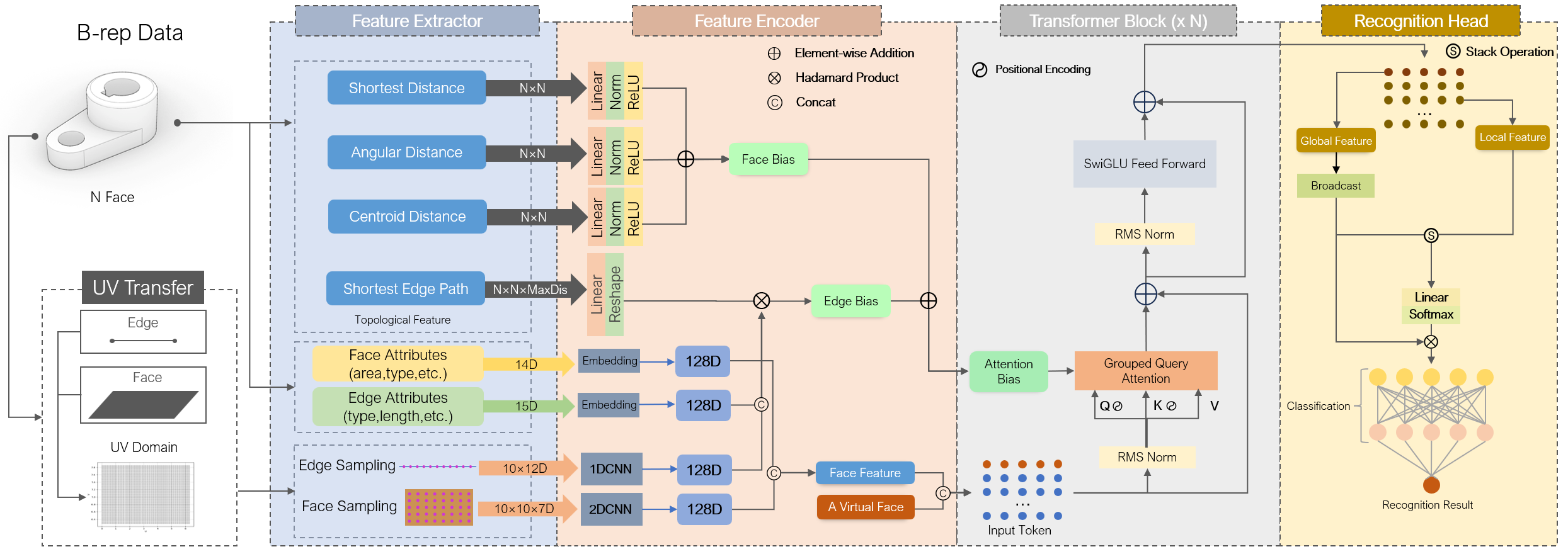} 
    \caption{
    % Our network accepts topological and geometrical features extracted from the B-rep model, with each set of features being encoded by a separate encoder. The edge features are combined with the topological features to serve as an attention bias, which is then fed into the designed transformer structure. The face features, after encoding, are augmented with a virtual face before being input into the transformer block. Finally, the two parts of features output by the transformer block are stacked together for feature fusion and then passed through a classification head to obtain the recognition results. 
    Our model consists of four main components: (1) Feature Extractor, which extracts topological and geometrical features from the B-rep model; (2) Feature Encoder, where edge and topological features are combined to create an attention bias and face features are augmented with a virtual face before being input into the transformer block; (3) Transformer Block, which processes the encoded features using grouped query attention and attention bias; and (4) Recognition Head, where the output features are fused and passed through a classification head to obtain the recognition results. 
    }
    \label{fig:Pipeline}
\end{figure*}

%-------------------------------------------------------------------------
\section{Related Work}
\label{sec:ReWork}

\subsection{Rule-Based Geometric Feature Recognition}
% Rule-based machining feature recognition methods rely on predefined rules to parse the topological and geometric information in CAD models to identify specific machining features. These rules, often based on engineering experience and industry standards, range from simple shape matching to complex combined feature recognition. However, comprehensively encoding all knowledge and rules related to machining features into the system is extremely challenging. This not only increases the system's complexity but also makes its maintenance and expansion cumbersome. 

% Traditional feature recognition algorithms primarily depend on the model's topological and geometric information. Early studies by Henderson \cite{henderson1984extraction}, Donaldson and Corney \cite{donaldson1993rule}, Chan and Case \cite{chan1994process} proposed rule-based methods that define boundary patterns and use expert systems to identify machining features. Despite their popularity due to intuitiveness and ease of implementation, these methods suffer from inherent limitations. The lack of uniqueness and completeness in rule definition results in poor system flexibility, especially when dealing with complex intersecting features. 

Rule-based geometric feature recognition methods identify geometric features in CAD models by applying predefined rules based on topological and geometric information. Early studies by Henderson \cite{henderson1984extraction}, Donaldson \cite{donaldson1993rule}, and Chan \cite{chan1994process} introduced rule-based approaches that define boundary patterns and use expert systems for feature recognition. 
% While intuitive and easy to implement, these methods struggle with flexibility due to the inherent limitations of predefined rules. Encoding all machining knowledge into a rule-based system is challenging, making maintenance cumbersome and leading to poor adaptability, particularly in handling complex intersecting features. 
However, these methods lack flexibility due to the limitations of predefined rules, making it difficult to encode all machining knowledge and adapt to complex features.

% Volumetric decomposition is another widely recognized rule-based automatic feature recognition (AFR) method. It decomposes the removed volume in the CAD model into intermediate volumes and reconstructs machining features according to predefined rules. Woo et al. \cite{woo1982feature} proposed the Alternating Sum of Volumes (ASV) decomposition method, representing the solid as the difference and union of convex bodies in a tree structure. Requicha et al. \cite{vandenbrande1993spatial} developed an AFR method that decomposes the volume to be machined into manufacturable features and handles feature interactions through generation-testing strategies and computational geometry techniques. Wilde et al. \cite{kim1992convergent} improved the ASV method by introducing partitioning, solving the convergence issue and exploring its application in feature recognition.

Volumetric decomposition is another well-established rule-based approach for geometric feature recognition. This method decomposes the material to be removed from a CAD model into intermediate volumes and reconstructs geometric features based on predefined rules. Woo et al. \cite{woo1982feature} introduced the Alternating Sum of Volumes (ASV) method, representing solids as hierarchical convex bodies. Requicha et al. \cite{vandenbrande1993spatial} further developed this approach to decompose machinable volumes into manufacturable features, addressing interactions through generation-testing strategies. Wilde et al. \cite{kim1992convergent} refined the ASV method by introducing partitioning techniques to enhance its applicability.  

Graph-based methods have gained attention due to their alignment with the B-rep structure of CAD models. These methods construct an Attribute Adjacency Graph (AAG) to capture topological and geometric information, and identify features by detecting subgraph patterns. 
Joshi et al. \cite{joshi1988graph} first applied the AAG for topological and geometric information matching. Shah et al. \cite{gao1998automatic} enhanced this approach by integrating hint-based feature recognition, introducing minimal condition subgraphs to improve the handling of feature interactions. 

Hint-based methods \cite{han1997integration, li2015hint} use rules to identify complex intersecting features by extracting patterns and heuristic "hints" from residuals left by feature intersections. 
These hints guide the reconstruction of incomplete feature information through reasoning. 
While this method improves the recognition of intersecting features and aids subsequent process planning, defining comprehensive and precise hints, along with robust reasoning rules, remains challenging. 
This complexity makes the hint-based method challenging to achieve fully automated and high-precision feature recognition in practical applications. 
% The structural disruptions caused by feature intersections complicate the process, limiting the feasibility of fully automated and high-precision feature recognition in practical applications.

\subsection{Deep Learning-Based Geometric Feature Recognition}

Various deep learning-based methods \cite{lei2022mfpointnet,yao2023machining, xu2024retrieval,huang2024frozen,huang2025psreg} have been developed for 3D structural representation, each addressing different challenges. Point-cloud-based methods leverage neural networks to extract features but often suffer from information loss. MFPointNet \cite{lei2022mfpointnet} employs selective downsampling layers for feature recognition, while Yao et al. \cite{yao2023machining} proposed a hierarchical neural network to improve the recognition of complex overlapping features. Shi et al. \cite{shi2020novel} introduced a multi-sectional view (MSV) representation and MsvNet, enriching 3D model representation by incorporating multi-view features. 

% Voxel-based approaches use 3D convolutional neural networks (CNNs) to process CAD models but face resolution-related information loss. FeatureNet \cite{zhang2018featurenet} applies 3D CNNs for feature recognition, while Peddireddy et al. \cite{peddireddy2020deep, peddireddy2021identifying} refined voxelization techniques to predict machining processes like milling and turning. Despite these improvements, voxelization inherently reduces geometric fidelity, particularly at low resolutions. Mesh-based approaches seek to retain geometric details for improved recognition. Jia et al. \cite{jia2023machining} proposed an innovative method that combines the original MeshCNN with Faster RCNN, forming a geometric feature recognition scheme based on Mesh Faster RCNN. This approach enhances the accuracy of geometric feature detection while preserving the mesh geometry. However, the high memory demand when processing high-resolution data limits its application in large-scale scenarios. 

Voxel-based approaches use 3D CNNs to process CAD models but suffer from resolution-related information loss. For example, FeatureNet \cite{zhang2018featurenet} applies 3D CNNs for feature recognition, and Peddireddy et al. \cite{peddireddy2020deep, peddireddy2021identifying} refined voxelization techniques to predict machining processes like milling and turning. However, voxelization inherently reduces geometric fidelity, especially at low resolutions. 
Mesh-based methods aim to retain geometric details for better recognition. Jia et al. \cite{jia2023machining} proposed a method combining MeshCNN with Faster RCNN (Mesh Faster RCNN) to enhance detection accuracy while preserving mesh geometry. However, the high memory demand for processing high-resolution data limits its scalability in large-scale scenarios.

In the CAD industry, B-rep is the dominant format for 3D models, making graph-based representations particularly effective. Graph neural networks (GNNs) have been widely applied due to their structural similarity with B-rep. For example, Cao et al. \cite{cao2020graph} pioneered the transformation of B-rep models into graph-structured representations for learning. Colligan et al. \cite{colligan2022hierarchical} introduced Hierarchical CADNet, a novel approach that leverages a two-level graph representation to improve recognition accuracy. Specifically, it utilizes the Face Adjacency Graph (FAG) to capture topological information and mesh patches to represent geometric details. Jayaraman et al. \cite{jayaraman2021uv} proposed UV-Net, which encodes surfaces and curves with CNNs and utilizes graph convolutional networks (GCNs) for feature learning.

Recent advancements further refine B-rep-based learning methods. Lambourne et al. \cite{lambourne2021brepnet} developed BRepNet, which defines convolution kernels for directed coedges, improving pattern detection. Lee et al. \cite{lee2023brepgat} introduced BRepGAT, incorporating graph attention networks (GATs) for precise feature segmentation. Wu et al. \cite{wu2024aagnet} proposed AAGNet, a multi-task GNN that simultaneously performs semantic, instance, and base segmentation using the geometric attribute adjacency graph (gAAG). Xia et al. \cite{MFTRCAD} developed MFTReNet, which learns semantic segmentation, instance grouping, and topological relationship prediction directly from B-rep data. Despite these advancements, the generation of large-scale datasets with detailed topological labels requires substantial annotation efforts, thereby increasing the cost of data preparation. 

% one major challenge remains -- generating large-scale datasets with detailed topological labels requires extensive manual annotation, increasing data preparation costs and limiting the scalability of deep learning models for machining feature recognition.

%-------------------------------------------------------------------------

\section{Method}
\label{sec:Method}
\subsection{Overview}  % 直接Overview就行 -- 应该强调某几个technical contribution, 然后写一下motivation
% BRepTransformer features in xxx, which could improve xxx

We propose a novel approach for CAD geometric feature recognition based on a transformer architecture as show in Figure~\ref{fig:Pipeline}, consisting of four different parts : 1) \textbf{Feature Extractor Module} considers the topological and geometric features of the model from multiple perspectives; 2) ~\textbf{Feature Encoder Module} encodes the initial input features into a format that is friendly for the network, thereby generating the main integrated features; 3) \textbf{Transformer Block Module} further extracts features using a transformer structure, with a designed attention bias constraint; 4) \textbf{Recognition Head Module} fuses the features and classifiers the faces of B-rep data.

\subsection{Feature Extraction Module}
\label{featureExtractionModel}
The Feature Extraction Module extracts feature for both the geometry and topology of the B-rep model, focusing on faces and edges. 

\subsubsection{Topological Feature Extraction}
For topology, we extract four different features, including three face features (Face Shortest Distance, Face Angular Distance and Face Centroid Distance) and one feature for edges (Shortest Edge Path). 

\textbf{Face Shortest Distance}. To fully capture the direct and indirect spatial relationships between any two faces in the B-rep model, we employed the Dijkstra algorithm to compute the shortest path length between all pairs of faces. This method quantifies the topological distance between faces as the number of intervening faces along the connecting path. Based on these results, we constructed an extended adjacency matrix $M_{d}\in R^{N\times N}$, where each element $m_d(f_i,f_j)$ represents the shortest distance from any face $f_i$ to another face $f_j$. This matrix captures not only direct connections between faces but also indirect connections via other faces, thereby reflecting complex connectivity patterns. 

\textbf{Face Angular Distance}. To accurately describe the relative position between two faces, our approach extracts the dihedral angle between them. For adjacent faces sharing a common edge, the dihedral angle indicates their geometric relationship (e.g., concave, convex, or smooth transition). For non-adjacent faces, we calculate the angle between their normal vectors, reflecting their relative orientation in 3D space regardless of intermediate faces. At the same time, our approach also constructs a matrix $M_a \in \mathbb{R}^{N \times N}$ to represent the angular information between faces, where each element $m_a(f_i,f_j)$ represents the angle between a face $f_i$ and another face $f_j$. 

\textbf{Face Centroid Distance}. The centroid distance between faces, a key topological feature, quantifies their spatial separation by measuring the Euclidean distance between face centroids. To eliminate scale influence, we normalize this distance by the bounding box diagonal length of the solid model, yielding a relative distance indicator. This normalized distance reflects similarity between CAD models of varying sizes and proportions. Similarly, we construct a matrix $M_c \in \mathbb{R}^{N \times N}$ to represent the angular information between faces, where each element $m_c(f_i,f_j)$ represents the Euclidean distance between the centroids of face $f_i$ and face $f_j$. 

\textbf{Shortest Edge Path}. Considering the key role of edges in defining global topological features, our approach particularly focuses on the shortest edge path between any two faces. Edges are not only the basic elements connecting different faces but also carry rich information about the model's internal connectivity and surface continuity. Therefore, for any two face $f_i$ and $f_j$ in the B-rep model, we not only calculate the shortest distance between them but also record all the edge chains $\left\{e_{i k}, e_{k l}, \ldots, e_{m j}\right\}$ that make up this shortest path. To effectively represent this edge path information, we introduce a three-dimensional matrix $M_e \in \mathbb{R}^{N \times N \times \text {MaxDistance}}$ to store edge path data. The parameter MaxDistance is employed to specify the upper limit of the distance between any two surfaces within a single model. For any pair of surfaces $\left(f_i, f_j\right)$, in a given model, when the shortest path distance between them is less than MaxDistance, the elements in the edge path matrix that exceed the actual shortest path range will be initialized to -1 as an indicator of invalid values.

\subsubsection{Geometric Feature Extraction}
Our approach further conducts geometric feature extraction from two aspects: the UV domain and geometric attributes, as detailed below. 

\begin{figure}[t]
    \centering
    \includegraphics[width=\columnwidth]{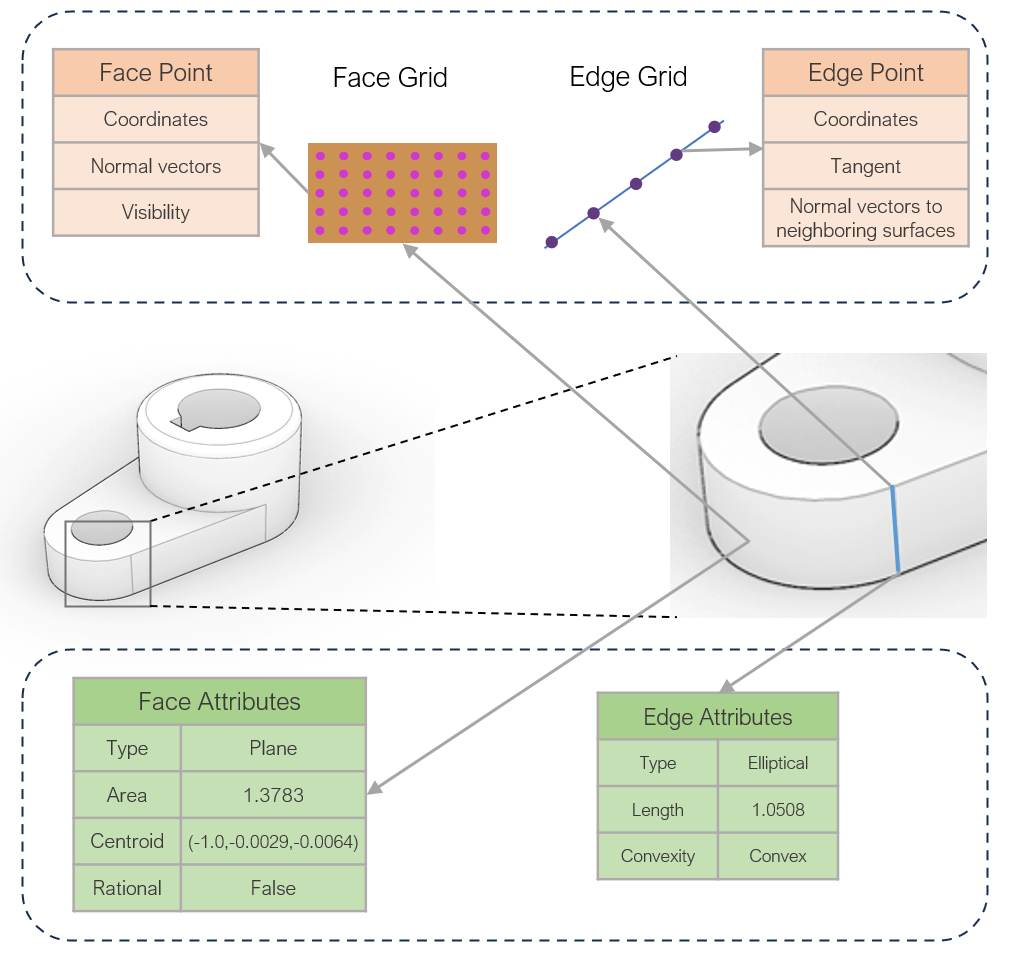} 
    \caption{ The upper part of the figure shows the details of geometric UV domain sampling, while the lower part shows the details of geometric attribute sampling. }
    \label{fig:UV-sampling}

\end{figure}

For extracting UV domain features from CAD models, we follow a method inspired by UV-Net \cite{jayaraman2021uv}. We sample and discretize the parametric surfaces and parametric curves surfaces in B-rep into regular 2D and 1D point grids with a uniform step size, as shown in Figure~\ref{fig:UV-sampling}. Each face's grid includes 3D coordinates, normal vectors of the points, and a visibility indicator. Each edge's grid includes 3D coordinates, tangent vectors, and the normal vectors of the adjacent faces. Compared to discrete methods like voxel grids \cite{yan2023manufacturing} and traditional meshes \cite{feng2019meshnet}, our UV grids capture precise geometric details and provide a suitable input format for neural networks. 
% Figure \ref{fig:UV-sampling} illustrates the sampling points for faces and edges in our approach. 

Furthermore, our method also extracts the geometric attributes inherent in the solid entities of CAD models as show in Figure~\ref{fig:UV-sampling}. To characterize the geometric attribute features of surfaces, we extract the following information: surface type (e.g., plane, conical surface, cylindrical surface, etc.), area, centroid coordinates, and whether it is a rational B-spline surface. For the geometric attribute features of curved edges, we extract their type (e.g., circular, closed curve, elliptical, straight line, etc.), length, and convexity (i.e., concave, convex, or smooth transition). These geometric attributes of faces and edges can be directly obtained from the original B-rep structure and are encoded separately. By integrating the aforementioned geometric attributes extracted from the UV domain, we obtain a geometric input representation of the entire CAD model, which provides strong support for subsequent downstream tasks. %Figure \ref{fig:UV-sampling} illustrates the details of our geometric feature sampling. 

% -------------------------------
\subsection{Feature Encoder Module}
\label{featureEncoderModel}
Feature encoder module further encodes all features extracted above, and output an attention bias and faces features, as the input for the following module. 

\subsubsection{Topological Feature Encoder.} 
For the above-mentioned three topological relation matrices $M_d, M_a, M_c \in \mathbb{R}^{N \times N}$ between faces, we applied a unified processing pipeline to these matrices. Specifically, each matrix is first transformed through a linear layer of the same dimension, followed by sequential processing through a normalization layer and a ReLU activation function for feature encoding, which is formulated as: 

\begin{equation}
M_i' = \text{ReLU}(\text{Norm}(\text{Linear}((M_i)))
\end{equation}
here, $M_i$ denotes the three topological matrices $M_d, M_a, \text{and}\ M_c$. 

% For the shortest distance matrix $M_d \in \mathbb{R}^{N \times N}$, 
% %in the experimental dataset, the shortest distance between any two faces does not exceed 256 units. Therefore, 
% our approach employs a linear layer of dimensions (256, 32) to encode the distance features between any two faces, followed by a normalization layer and a ReLU activation function:

% \begin{equation}
% h_{M_d}\ = \text{Embedding}\{M_d, (256, 32)\} \in \mathbb{R}^{N \times N \times 32}
% \end{equation}

% For the centroid distance matrix $M_c$ and the dihedral angle distance matrix $M_a$, the two matrices extract uncertain scalar values. Therefore, we employed two Linear layers with identical structures to encode these two pieces of information into the same feature representation as mentioned above. 

Based on these three parts, we obtain the face bias input into the transformer module: 
\begin{equation}
B_{\text{Face}} = \text{Add}({M_d'}, {M_a'}, {M_c'})
\end{equation}
% where Add operation is xxxx, 
within the framework of the edge path matrix, we have considered the impact of edge weights on faces. For the shortest path between any two faces in a B-rep model, the influence of distant edges on the initial face diminishes progressively as the position along the path advances from the starting face. Based on these considerations, our method integrates the edge path matrix with edge features to model the weight influence of edges on faces along the path. The formula is expressed as follows: 
% Regarding the edge path matrix, considering the influence of Edge on node, as the position on the path progresses, the influence of the edge on the starting node gradually decreases. Therefore, a learnable weight $w$ is introduced in our approach to model the influence of Edge on node, where the longer the path, the less important the newly added edge becomes to the original node:

\begin{equation}
B_{Edge}=\ {(M}_e\ \otimes H_{edge})
\end{equation}
here, $M_e$ stands for the edge path matrix, and $H_{edge}$ indicates the extracted and encoded edge features. 

Finally, the sum of the obtained $B_{Edge}$ and $B_{Face}$ yields the attention bias that we input into the transformer module as shown in Figure~\ref{fig:Pipeline}.

\subsubsection{Geometric Feature Encoder} 
For the encoding of UV domain sampling features in CAD geometric features, \systemname~ encodes face feature as $f_{geo} \in \mathbb{R}^{N_f \times N_u \times N_v \times 7}$ , and edge feature as $e_{\text {geo}} \in \mathbb{R}^{N_e \times N_u \times 12}$. Here, $N_f$ and $N_v$ indicate the number of sampling points for faces and edges based on the UV grid parameters respectively. $N_u$ and $N_v$ represent the number of sampling points along the u-axis and v-axis. The aforementioned geometric features are all fed into their respective encoders. The face encoder consists of three 2D CNN layers, an adaptive average pooling layer, and a fully connected layer, which encode the face features into a feature dimension of 128. Similarly, the edge encoder has a structure akin to the face encoder but uses 1D CNN layers for preliminary encoding. The encoded information of geometric features is shown in Table \ref{tab:Geometric}.

\begin{table}[t]
\centering
\caption{Geometric Feature Dimensions in UV Domain}
\label{tab:Geometric}
\resizebox{\columnwidth}{!}{%
\fontsize{9}{12}\selectfont % 设置字体大小为9pt，行距为12pt
\begin{tabular}{lll}
\toprule
\textbf{Element} & \textbf{Feature} & \textbf{Dimension} \\
\midrule
Face & Coordinates & 3D \\
     & Normal vectors & 3D \\
     & Visibility & 1D \\
Edge & Coordinates & 3D \\
     & Tangent & 3D \\
     & Normal vectors to neighboring surfaces & 6D \\
\bottomrule
\end{tabular}
}
\end{table}

We employed Multilayer Perceptrons (MLP) to encode the geometric attribute features of faces and edges. The encoded face attributes include a 9-dimensional one-hot vector $ f_{\text{type}} \in \mathbb{R}^9 $ to represent surface types (e.g., plane, conical surface, cylindrical surface, etc.), a vector $ f_{\text{area}} \in \mathbb{R}^1 $ to identify the surface area, a vector $ f_{\text{cen}} \in \mathbb{R}^3 $ to identify the centroid coordinates of the face, and a vector $ f_{\text{rat}} \in \mathbb{R}^1 $ to identify whether it is a rational B-spline surface. The edge attributes include an 11-dimensional one-hot vector $ e_{\text{type}} \in \mathbb{R}^{11} $ for identifying edge types (e.g., circular, closed curve, elliptical, straight line, etc.), a vector $ e_{\text{len}} \in \mathbb{R}^1 $ to identify the type of the edge, and a 3-dimensional one-hot vector $ e_{\text{conv}} \in \mathbb{R}^3 $ to characterize the convexity of the edge (concave, convex, or smooth). The encoded information of attribute features is shown in Table \ref{tab:Attribute} as indicated.

\begin{table}[t]
\centering
\caption{Attribute Feature Encoding for Geometric Elements}
\label{tab:Attribute}
\resizebox{\columnwidth}{!}{%
\begin{tabular}{llll}
\toprule
\textbf{Element} & \textbf{Input Feature} & \textbf{Encoder Layer} & \textbf{Output} \\
\midrule
Face & $ f_{\text{type}} \in R^9 $ & Linear (9, 32) & $ h_{f,\text{type}} \in R^{32} $ \\
     & $ f_{\text{area}} \in R^1 $ & Linear (1, 32) & $ h_{f,\text{area}} \in R^{32} $ \\
     & $ f_{\text{cen}} \in R^3 $ & Linear (3, 32) & $ h_{f,\text{cen}} \in R^{32} $ \\
     & $ f_{\text{rat}} \in R^1 $ & Linear (1, 32) & $ h_{f,\text{rat}} \in R^{32} $ \\
Edge & $ e_{\text{type}} \in R^{11} $ & Linear (11, 64) & $ h_{e,\text{type}} \in R^{64} $ \\
     & $ e_{\text{len}} \in R^1 $ & Linear (1, 32) & $ h_{e,\text{len}} \in R^{32} $ \\
     & $ e_{\text{conv}} \in R^3 $ & Linear (3, 32) & $ h_{e,\text{conv}} \in R^{32} $ \\
\bottomrule
\end{tabular}
}
\end{table}

Finally, by concatenating the encoded features of the two types of geometric attributes and the geometric UV domain features, we obtained the complete geometric face and edge features, $H_{\text{face}}$ and $H_{\text{edge}}$ respectively:

\begin{equation}
H_{\text{face}} = \text{Concat}(h_{f,\text{geo}}, h_{f,\text{type}}, h_{f,\text{area}}, h_{f,\text{cen}}, h_{f,\text{rat}}) \in \mathbb{R}^{256}
\end{equation}

\begin{equation}
H_{\text{edge}} = \text{Concat}(h_{e,\text{geo}}, h_{e,\text{type}}, h_{e,\text{len}}, h_{e,\text{conv}}) \in \mathbb{R}^{256}
\end{equation}

\begin{figure*}[ht]
    \centering
    \includegraphics[width=\textwidth]{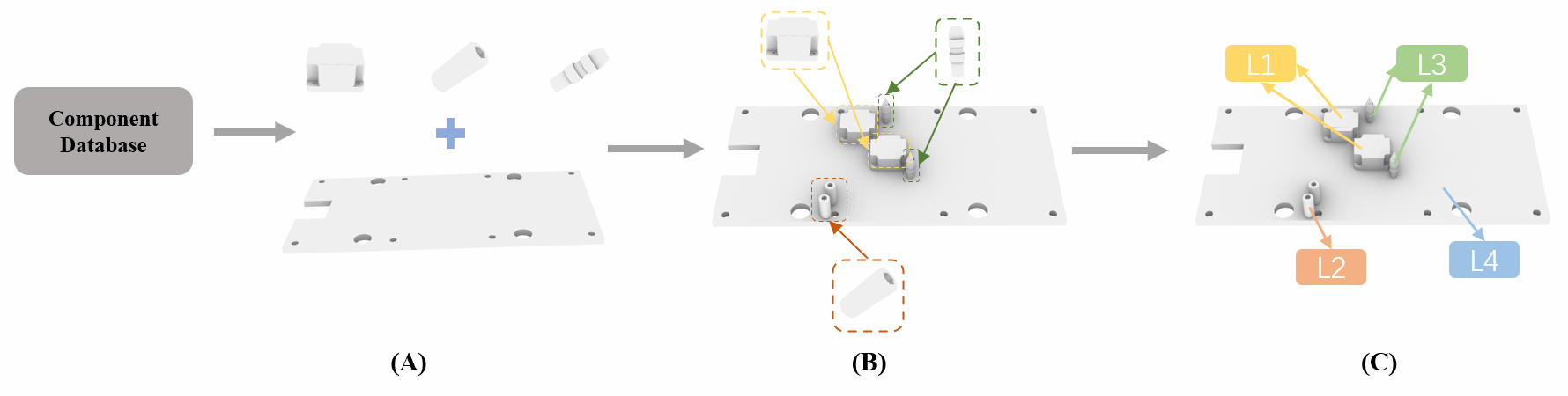} 
    \caption{The entire process of constructing our dataset, (A) illustrates the selection of four components from the database, 
    % (B) denotes the operation of random rotation, translation, and scaling based on geometric constraints,
    (B) shows the application of random rotation, translation, and duplication under geometric constraints to generate a new B-rep model, 
    and (C) demonstrates the process of traversing and labeling all faces (L1, L2, L3, L4). }
    \label{fig:creating}

\end{figure*}

\subsection{Transformer Block Module}
\label{transformerModule}
%Most existing methods for feature recognition in B-rep models are based on traditional Graph Neural Networks (GNNs) \cite{MFTRCAD}, particularly Message Passing Neural Networks (MPNNs) \cite{wu2024aagnet}. Although these methods have achieved certain results, they are typically constrained by the information propagation mechanism within local neighborhoods, resulting in limited receptive fields and an inability to capture global context effectively. Moreover, existing approaches often overlook the importance of edge features, failing to fully utilize all available information within the B-rep structure. 
% Our transformer module is a novel Transformer framework tailored for processing input data, and concurrently, we have specifically devised an Attention bias mechanism to augment the flexibility and specificity of feature propagation. Our approach not only transcends the confines of local neighborhoods, enabling each face to interact with any other face within the B-rep, but also effectively amalgamates edge features and global topological features through meticulously designed attention biases, thereby more adeptly capturing the intricate topological and geometric relationships inherent in complex CAD models. 

Before feeding the encoded face features into our transformer architecture, we introduced a virtual face feature that is connected to all B-rep elements. This virtual face feature, within the transformer structure, can engage in deep interactions with the actual face features, thereby obtaining a global feature that represents the entire B-rep model. 

This part consists of 8 layers of designed transformer modules, with each layer primarily incorporating Grouped Query Attention (GQA)~\cite{ainslie2023gqa}, Root Mean Square Normalization(RMS Norm), and SwiGLU activation function. First, all the feature of the faces are represented as $H_{\text{face}}^0 = \left[ {h_{\text{face}}^1} , {h_{\text{face}}^2}, \ldots, {h_{\text{face}}^{N+1}} \right] \in \mathbb{R}^{{(N+1)} \times 256}$. Here, each $h_{\text{face}}^{i}$ represents the feature of a given face, $N + 1$ indicates that it includes the features of the virtual face, and the superscript 0 in $H_{\text{face}}^0$ denotes the initial iteration of the overall feature representation. 

% \begin{equation}
% \bar{a}_i = \frac{a_i}{\text{RMS}(a)} g_i \quad \text{where} \quad \text{RMS}(a) = \sqrt{\frac{1}{n} \sum_{i=1}^{n} a_i^2}
% \end{equation}

First, all the input face features are fed into RMS Norm for normalization. Next, the normalized features are fed into the GQA for feature propagation. Following this, the features processed by the attention mechanism are fed into the Feed-Forward Network (FFN), and then combined with residual connections for further processing. The overall formula expressed as follows: 

\begin{equation}
H_{\text{face}}^t{}' = \text{Attention}\left(\text{RMS Norm}(H_{\text{face}}^{t-1})\right) + H_{\text{face}}^{t-1}
\end{equation}

\begin{equation}
H_{\text{face}}^t = \text{FFN}\left(\text{RMS Norm}(H_{\text{face}}^{t}{}')\right) + H_{\text{face}}^{t}{}'
\end{equation}

Our attention uses GQA, which divides the queries into multiple groups and independently computes attention within each group. It then concatenates the outputs of all groups $Q_g$ and passes them through a linear transformation to obtain the output, effectively reducing computational load and memory usage, and $t$ indicates the iteration times. The specific expression formula for GQA is as follows: 

\begin{equation}
O = \left( \text{Concat} \left[ \text{softmax}\left( \frac{Q_g K^T}{\sqrt{d_k}} + B_{\text{Att}} \right) V \right]_{g=1}^{G} \right) W_o
\end{equation}
where $O$ refers to the final output, $G$ indicates the upper limit of the number of groups. For each group $g$, the dot product of $Q_g$ and $K$ is first scaled by the square root of the key vector dimension $d_k$. This scaled product is then adjusted by the attention bias $B_{att}$, and the $softmax$ function is applied to obtain the attention weights. 
These weights are used to compute the weighted sum of the values $V$. The outputs of all groups are concatenated and passed through a linear transformation $W_o$ to produce the final output $O$.

In the FFN layer, we primarily use the SwiGLU activation function, which is mathematically expressed as follows: 

\begin{equation}
\text{SwiGLU} = \text{Swish}(Wx + b) \otimes (Vx + c)
\end{equation}
here, $W$ and $V$ denote the learnable weight matrices that are applied to the input $x$, $b$ and $c$ are bias terms. The Swish activation function defined as $\text{Swish}(z) = z \cdot {sigmoid}(z) $. 

% Finally, to obtain the global features of the entire B-rep model, we introduce a global virtual face during the feature propagation process. This virtual face, acting as a surrogate entity, interacts with all other faces within the B-rep model during the network's information propagation. Through this approach, we derive a comprehensive global feature representation of the B-rep model, denoted as $f_{\text{global}} \in \mathbb{R}^{256}$.
Finally, based on the introduced global virtual face feature and the input encoded face features, the module outputs $f_{global}$ , representing the global feature of the B-rep model, and $f_{local}$ , representing the local feature.

% \begin{figure}[t]
%     \centering
%     \includegraphics[width=\columnwidth]{images/CBFCAD2.png}
%     \caption{Three types of distinct component objects and their random position on the base plate.}
%     \label{fig:Base Dataset}
% \end{figure}

\subsection{Recognition Head}
% Based on the obtained local feature representation $f_{local}$ of the B-rep and the global feature representation $f_{global}$ , it naturally occurs to us to fuse these two parts of features. Therefore, drawn from Attention \cite{vaswani2017attention}, our approach designs a feature fusion structure to perform feature integration, effectively combining $f_{global}$ with $f_{local}$. The specific formulas are as follows: 
Based on the local feature representation of $f_{local}$ and the global feature $f_{global}$ output by the transformer module, we design a structure to fuse these two parts of features. First, we broadcast the global feature to the same dimension as the local feature and stack the two parts along a new dimension to obtain the integrated feature representation $F_{all}$. Then, we multiply the features $F_{all}$ by a learnable weight matrix $W_w$ and apply a $softmax$ activation function to obtain the weight representation of these features. Finally, we perform element-wise multiplication between this weight representation and the $F_{all}$ to generate the final output features. The specific formulas are as follows:

\begin{equation}
F_{all} = \text{$stack$}[f_{local}, f_{global}]
\end{equation}

\begin{equation}
F_{\text{out}} = F_{\text{all}} \otimes \text{softmax}(F_{\text{all}} \cdot W_w + b_w)
\end{equation}
where $stack$ denotes the operation of concatenation along a new dimension, $F_{all}$ represents the newly obtained features, $W_w$ represents the learnable weight matrix of the linear layer and $b_w$ denotes the bias term associated with it. 

For the final feature recognition task in CAD models, we employed a MLP coupled with an $argmax$ activation function as the classifier to predict the geometric feature category $\hat{C}$ of each face. The mathematical expression is as follows: 

\begin{equation}
\hat{C} = \arg\max(\text{MLP}(F_{\text{out}}) \in \mathbb{R}^{N \times K})
\end{equation}

We choose the cross-entropy as our final loss function, which is as follow:

\begin{equation}
L = -\frac{1}{N} \sum_{i=1}^{N} \sum_{j=1}^{C} y_{i,j} \log (\hat{C}_{i,j} + \epsilon)
\end{equation}
here, $\epsilon$ is a small positive constant added to the predicted probabilities to ensure numerical stability

%我们推出了一个支持几何复杂特征识别研究的数据集，名为xxx，同时介绍了创建相应数据集的方法。该数据集一共有1500个B-rep数据格式的CAD模型。每个模型是一个母板和其上的三种不同几何特征通过布尔组合形成的B-rep模型。每个几何特征所包含的面都被打了相应的标签，标签信息保存在一个独立的json文件中，以支持对这些复杂特征的模型构建和训练的相关研究。图[]展示了母板和三类几何特征的形状，以及他们组合后的一个模型的整体外观。在制作数据过程中，我们先通过相关三维模型编辑软件，从工业数据中挑选数据并进行母板和几何特征的分离。然后，我们对提取出的三种几何特征在底板上进行了随机旋转、平移和缩放。在进行这些操作时，我们通过操作模型的几何约束来保证了生成模型的合理。具体来说，平移时，脚本沿着接触面的法向量和切线向量方向移动模型，直到模型与底座相交，确保部件与底座正确连接;而旋转和放缩则基于接触面的几何属性（如法向量和质心）进行随机变换，避免模型变形或脱离底座。最后在标签标注阶段，我们遍历了合成模型的所有面，确定每个面属于哪个原始部件（底座、部件A、部件B、部件C），并为每个面分配一个标签，这些信息被存储在字典中，其中键为面的编号索引，值为对应的部件标签。图1展示了我们的整个操作流程。

\section{Complex B-rep Feature (CBF) Dataset }
\label{datasetCBF}
We introduce the CBF dataset to support research on complex geometric feature recognition. It contains 20,000 B-rep CAD models, each formed by combining a base plate with three geometric features. The faces of these features are labeled and stored in JSON files for model construction and training research. 
% Figure \ref{fig:Base Dataset} illustrates the shapes of the base plate and the three types of geometric features, as well as the overall appearance of a model formed by their combination. 

For our data creation process, Figure~\ref{fig:creating} demonstrates the entire workflow. We first selected B-rep from public sources and separated the base plates and geometric features using relevant 3D modeling software. Subsequently, we applied random rotations, translations, and duplications to the three extracted geometric features on the base plate. During these operations, we ensured the rationality of the generated models by manipulating their geometric constraints. Specifically, for translation, the script moved the model along the normal and tangent vectors of the contact surface until it intersected with the base plate, ensuring proper attachment. Rotations were performed based on the geometric attributes of the contact surface (such as normal vectors and centroids) to avoid model distortion or detachment. Duplication was carried out via boolean operations to ensure that the newly generated components did not conflict with the base plate or other components. Finally, in the labeling stage, we traversed all faces of the composite model, determined which original component each face belonged to (base plate, feature A, feature B, feature C), and assigned a label to each face. This information was stored in a dictionary, with face indices as keys and corresponding component labels as values.

% \begin{figure}[ht]
%     \centering
%     \includegraphics[width=\columnwidth]{images/Dataset2.png}
%     \caption{Three different types of physical objects are randomly distributed on the base.}
%     \label{fig:Base Dataset2}
% \end{figure}

\section{Experiments}
\label{sec:Experiments}
\subsection{Experimental Environment}
We trained our network using a single NVIDIA 4090 GPU and PyTorch-Lightning v1.9.0, highlighting its lightweight nature. During training, we used the AdamW optimizer with an initial learning rate of 0.001, and parameters set to $\beta_1 = 0.9$, $\beta_2 = 0.999$, and $\epsilon = {1 \times 10^{-8}}$ for stability. 
We also employed the ReduceLROnPlateau \cite{al2022scheduling} learning rate scheduler to dynamically adjust the learning rate based on validation loss. A linear learning rate warm-up was implemented for the first 5000 steps to aid convergence. The batch size was set to 64, and training lasted up to 200 epochs. 

To evaluate the performance of our network, we used overall accuracy $A$, class accuracy $A_c$, and mean Intersection over Union (mIoU) to verify the performance of our model's geometric feature recognition capabilities, mathematically expressed as follows:

\vspace{-5pt}

\begin{equation}
A = \frac{F_{\text{c}}}{F_{\text{t}}}
\end{equation}
here, $F_c$ denotes the number of correctly classified B-rep faces, and $F_t$ is the total number of B-rep faces in the CAD model. Overall accuracy A calculates the proportion of correctly classified B-rep faces to the total number of B-rep faces in the CAD model.

\begin{equation}
A_c = \frac{1}{C} \sum_{c=1}^{C} \frac{\hat{y}_c = l_c}{y_c = l_c}
\end{equation}
here, $A_c$ denotes the average accuracy across all classes. $y_c$ represents the total number of B-rep faces in label $l_c$, while $\hat{y}_c$ indicates the number of correctly predicted B-rep faces in label $l_c$. The class accuracy $A_c$ represents the average accuracy across all geometric feature classes.

\begin{equation}
mIoU = \frac{1}{C} \sum_{c=1}^{C} \frac{\hat{y}_c = l_c \cap y_c = l_c}{\hat{y}_c = l_c \cup y_c = l_c}
\end{equation}
here, the parameter expressions in this part are the same as those described above for $A_c$. The mIoU is commonly used to evaluate the overlap between predicted results and true labels. In the context of our geometric feature recognition task, mIoU is calculated as the average ratio of correctly classified B-rep faces that share the same class labels in both the actual and predicted outputs.

\subsection{Experimental Datasets}

% We have validated the performance of our model on the MFInstSeg and MFTRCAD public datasets and compared it with existing deep learning methods for geometric feature recognition, including UV-Net \cite{jayaraman2021uv}, Hierarchical CADNet \cite{colligan2022hierarchical}, AAGNet \cite{wu2024aagnet}, and others. The experimental results show that our model achieved state-of-the-art (sota) accuracy on both datasets. Additionally, we conducted comprehensive experiments on our proposed CBF dataset and compared our model with other open-source models. Our model also demonstrated sota performance in these experiments.
We evaluated our model's machining feature recognition ability on the MFInstSeg and MFTRCAD public datasets and compared it with mainstream deep learning methods. Our model achieved state-of-the-art accuracy on these datasets. We also tested its complex feature recognition capability on the proposed CBF dataset, where it again achieved state-of-the-art accuracy. For all datasets, we used a 70\% / 15\% / 15\% split for training, validation, and testing.

\subsubsection{MFInstSeg Dataset}
The MFInstSeg dataset comprises 62,495 CAD model files stored in B-rep format. It includes 24 different types of machining features, with each model containing 3 to 10 unique machining features. 
Table \ref{tab:MFInstSeg} shows the performance of our model and other mainstream models on this dataset.

\begin{table}[h]
\centering

\caption{Recognition Performance on MFInstSeg Dataset}
\label{tab:MFInstSeg}

\newcolumntype{Y}{>{\centering\arraybackslash}X}
\newcolumntype{Z}{>{\raggedright\arraybackslash}X}

\begin{tabularx}{\linewidth}{ZYY} 
\toprule
\textbf{Network} & \textbf{Accuracy(\%)} & \textbf{mIOU(\%)} \\
\midrule
ASIN\cite{zhang2022intelligent}  & $86.46 \pm 0.45$                      & $79.15 \pm 0.82$ \\
GATv2\cite{brody2021attentive}    & $95.90 \pm 0.20$                      & $93.03 \pm 0.36$ \\
GraphSAGE\cite{hamilton2017inductive}    & $97.69 \pm 0.06$                      & $95.70 \pm 0.14$ \\
GIN\cite{xu2018powerful}    & $98.14 \pm 0.03$                      & $96.52 \pm 0.06$ \\
DeeperGCN\cite{li2020deepergcn}    & $99.03 \pm 0.02$                      & $98.31 \pm 0.01$ \\
AAGNet\cite{wu2024aagnet}           & $99.15 \pm 0.03$                      & $98.45 \pm 0.04$ \\
MFTRNet\cite{MFTRCAD}          & $99.56 \pm 0.02$                      & $98.43 \pm 0.03$ \\
\textbf{\systemname(Ours)}   & $\textbf{99.62} \pm \textbf{0.03}$   &  $\textbf{98.74} \pm \textbf{0.09}$  \\

\bottomrule
\end{tabularx}

\end{table}

\vspace{-10pt}

\subsubsection{MFTRCAD Dataset}

The MFTRCAD dataset comprises 28,661 CAD models stored in B-rep format. The authors further divided one of the traditional 24 machining feature categories into three subcategories, leading to a total of 26 distinct machining features in the dataset. 
Table \ref{tab:MFTRCAD} presents the performance of our model and other mainstream models on this dataset.

\begin{table}[h]
\centering
\caption{Recognition Performance on MFTRCAD Dataset}
\label{tab:MFTRCAD}

\newcolumntype{Y}{>{\centering\arraybackslash}X}
\newcolumntype{Z}{>{\raggedright\arraybackslash}X}

\begin{tabularx}{\linewidth}{ZY} 
\toprule
\textbf{Network} & \textbf{Accuracy(\%)} \\
\midrule
PointNet++\cite{qi2017pointnet++}       & $67.89 \pm 0.08$ \\
DGCNN\cite{wang2019dynamic}            & $67.97 \pm 0.07$ \\
ASIN\cite{zhang2022intelligent}             & $68.57 \pm 0.41$ \\
Hierarchical CADNet\cite{colligan2022hierarchical} & $78.39 \pm 0.03$ \\
AAGNet\cite{wu2024aagnet}           & $79.45 \pm 0.02$ \\
MFTRNet\cite{MFTRCAD}          & $89.88 \pm 0.02$ \\
\textbf{\systemname(Ours)}               & $\textbf{93.16} \pm \textbf{0.11}$ \\
\bottomrule

\end{tabularx}

\end{table}

\vspace{-10pt}

\subsubsection{Complex Feature Dataset}

In our CBF dataset, unlike previous datasets, this dataset requires the model to identify these three distinct geometric features along with the base plate. 
The experimental results of our model and other mainstream models on this dataset is presented in Table \ref{tab:CBF}. 
% indicate that our model outperforms MFReNet, thereby showcasing its significant advantages.

\begin{table}[h]
\centering
\caption{Recognition Performance on CBF Dataset}
\label{tab:CBF}
\resizebox{\columnwidth}{!}{%
\begin{tabular}{lccc}
\toprule
\textbf{Network} & \textbf{Accuracy(\%)} & \textbf{Class Acc(\%)} & \textbf{mIoU(\%)} \\
\midrule

AAGNet\cite{wu2024aagnet} & 93.41  & -  & \textbf{93.76}   \\
MFTRNet\cite{MFTRCAD} & 92.12  & -  & 91.21   \\
\textbf{\systemname}(Ours) & \textbf{94.66}  & \textbf{94.97}  & 87.48    \\
% - & -  & -  & -   \\

\bottomrule
\end{tabular}
}
\end{table}

Although our network outperforms other comparative networks in terms of overall accuracy, it performs poorly in the mIoU metric. Analysis reveals that when the network identifies simple geometric features, the limited number of faces involved means that any misidentification significantly impacts overall accuracy. In contrast, when dealing with complex geometric features, the higher number of faces means that misidentification of some faces has a relatively limited impact on overall accuracy. Based on the above analysis, despite certain shortcomings in the mIoU metric, our network still maintains a leading position in overall recognition accuracy. This indicates that our network is more adept at recognizing complex, multi-faceted geometric features.

\subsection{Ablation Study}

In the ablation study, we focused on the impact of the input features of our model on its performance. We systematically removed these key features and tested the changes in model performance. All ablation experiments were conducted on our proposed CBF dataset. 

\subsubsection{Ablation Analysis of Geometric Features}

% The geometric features directly input into our \systemname~ network include the UV domain geometric features and the attribute features of the B-rep. In this ablation study, a model with complete input features was used as the baseline, and then each of the three input features was removed one by one to generate the ablation models. The results shown in Table \ref{tab:Ablation Analysis of Geometric Features} indicate that the removal of any input feature leads to a decrease in feature recognition accuracy. Among them, the removal of attribute features causes the most significant performance drop, while the removal of UV domain geometric features has a less noticeable impact. This suggests that attribute features are more important than the input geometric features within our network architecture. 

In our ablation study, we tested the importance of different input features for the \systemname~network. The network takes UV domain geometric features and B-rep attribute features as inputs. We compared a baseline model with complete inputs to models that removed each of these features individually. The results in Table \ref{tab:Ablation Analysis of Geometric Features} show that removing any input feature reduces feature recognition accuracy. Specifically, removing attribute features causes the most significant performance drop, while removing UV domain geometric features has a smaller impact. This indicates that attribute features are more critical than geometric features in our network architecture.

\begin{table}[h]
\centering

\vspace{-10pt}

\caption{Impact of Removing Different Geometric Features on \systemname~ Performance}

\resizebox{\columnwidth}{!}{%
\begin{tabular}{lccc}
\toprule
\textbf{Input} & \textbf{Accuracy(\%)} & \textbf{Class Acc(\%)} & \textbf{mIoU(\%)} \\
\midrule
\textbf{Full (baseline)} & \textbf{94.66} & \textbf{94.97} & \textbf{87.48} \\
w/o Face Attr    & 92.34 (-2.32)    & 91.23 (-3.74)            & 84.10 (-3.38)       \\
w/o Edge Attr    & 91.01 (-3.65)    & 90.01 (-4.96)        & 81.69 (-5.79)    \\
w/o UV-grid      & 92.95 (-1.71)     & 91.85 (-3.12)     & 85.05 (-2.43)       \\   
\bottomrule
\end{tabular}
}
\label{tab:Ablation Analysis of Geometric Features}
\end{table}

\vspace{-10pt}

\subsubsection{Ablation Analysis of Topological Features}

In our {\systemname} network, we focused on extracting four key topological features from CAD models to delineate the comprehensive relational structure of B-rep models. To demonstrate the effectiveness of the extracted features, we established a baseline model using the initial complete topological features and then conducted ablation experiments by progressively removing topological features on our CBF dataset. As shown in Table \ref{tab:Topological}, the network's accuracy decreased successively with the removal of each of the topological features. This indicates that the global topological matrices facilitate accurate understanding of complex 3D solid models by neural network. 

\begin{table}[h]
\centering

\caption{Impact of Removing Different Topological Features on \systemname~ Performance}

\resizebox{\columnwidth}{!}{%
\begin{tabular}{lccc}
\toprule
\textbf{Input} & \textbf{Accuracy(\%)} & \textbf{Class Acc(\%)} & \textbf{mIoU(\%)} \\
\midrule
\textbf{Full (baseline)} & \textbf{94.66} & \textbf{94.97} & \textbf{87.48} \\
w/o  $M_d$ & 93.51 (-1.15) & 92.73 (-2.24) & 86.28 (-1.20) \\
w/o  $M_d$, $M_a$ and $M_c$ & 93.33 (-1.33) & 92.29 (-2.68) & 85.95 (-1.53) \\
w/o  $M_d$, $M_a$, $M_c$ and $M_e$ & 93.22 (-1.44) & 92.17 (-2.80) & 85.60 (-1.88) \\
\bottomrule
\end{tabular}
}
\label{tab:Topological}
\end{table}

\subsection{Geometric Recognition Presentation}
\begin{figure}[h]
    \centering
    \includegraphics[width=\columnwidth]{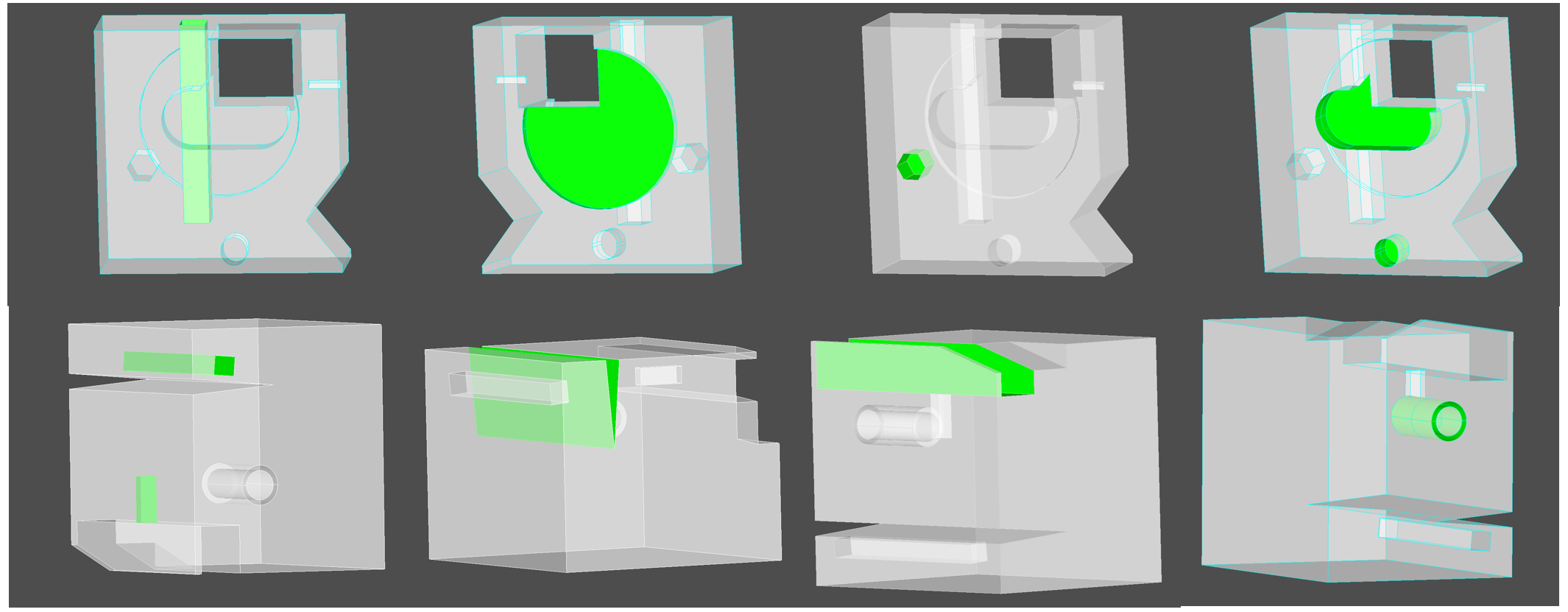} 
    \caption{Examples of recognized machining features (highlighted in green) in B-rep models}
    \label{fig:CAD}
\end{figure}

\begin{figure}[h]
    \centering
    \includegraphics[width=\columnwidth]{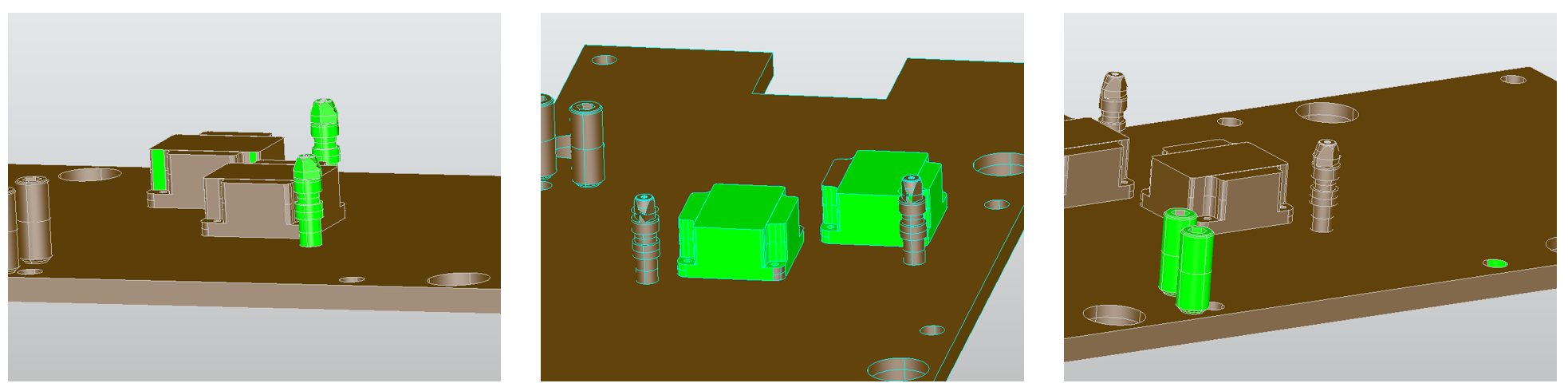} 
    \caption{Examples of recognized complex features (highlighted in green) in B-rep models}
    \label{fig:complex}
\end{figure}

% To describe some cases where the model fails to capture the feature.
This section presents a visual demonstration of our network's performance in geometric feature recognition. Figure~\ref{fig:CAD} illustrates the network's capability in identifying machining features, with the green highlights indicating the features that have been successfully recognized by the network. Figure~\ref{fig:complex}, meanwhile, displays the outcomes of our network's feature recognition on more complex geometric shapes.

\vspace{-20pt}

\vspace{-20pt}

\section{Conclusion}
In this paper, we introduce \systemname, a novel geometric feature recognition network based on the transformer architecture. Our network effectively extracts both geometric and topological information from CAD models, and incorporates an attention bias that integrates geometric and topological features to regulate information propagation within the transformer module. Furthermore, we propose the CBF dataset, which features more complex geometric and topological representations and is specifically designed for complex feature recognition tasks. Finally, \systemname~ achieves state-of-the-art accuracy on the public MFInstSeg and MFTRCAD datasets, as well as our CBF dataset, thereby demonstrating its superiority in both machining feature recognition and complex geometric feature recognition tasks.

\section{Acknowledgments}
This research was funded by the Fundamental Research
Funds for the Central Universities, the Natural Science Basic Research Program of Shaanxi Province under Grant
2024JC-YBQN-0702, and NUS Research Scholarship and
NUS ORIA.

\clearpage
\bibliographystyle{ACM-Reference-Format}
\bibliography{sample-base}

%%
%% If your work has an appendix, this is the place to put it.

% \clearpage
% \appendix

% \section{Research Methods}

% \subsection{Part One}

% Lorem ipsum dolor sit amet, consectetur adipiscing elit. Morbi
% malesuada, quam in pulvinar varius, metus nunc fermentum urna, id
% sollicitudin purus odio sit amet enim. Aliquam ullamcorper eu ipsum
% vel mollis. Curabitur quis dictum nisl. Phasellus vel semper risus, et
% lacinia dolor. Integer ultricies commodo sem nec semper.

% \subsection{Part Two}

% Etiam commodo feugiat nisl pulvinar pellentesque. Etiam auctor sodales
% ligula, non varius nibh pulvinar semper. Suspendisse nec lectus non
% ipsum convallis congue hendrerit vitae sapien. Donec at laoreet
% eros. Vivamus non purus placerat, scelerisque diam eu, cursus
% ante. Etiam aliquam tortor auctor efficitur mattis.

% \section{Online Resources}

% Nam id fermentum dui. Suspendisse sagittis tortor a nulla mollis, in
% pulvinar ex pretium. Sed interdum orci quis metus euismod, et sagittis
% enim maximus. Vestibulum gravida massa ut felis suscipit
% congue. Quisque mattis elit a risus ultrices commodo venenatis eget
% dui. Etiam sagittis eleifend elementum.

% Nam interdum magna at lectus dignissim, ac dignissim lorem
% rhoncus. Maecenas eu arcu ac neque placerat aliquam. Nunc pulvinar
% massa et mattis lacinia.

\end{document}